\documentclass[letterpaper, 10 pt, conference]{ieeeconf}

\IEEEoverridecommandlockouts
\usepackage{cite}
\usepackage{amsmath,amssymb,amsfonts}
\usepackage{algorithm}
\usepackage{algpseudocode}
\usepackage{graphicx}
\usepackage{booktabs}
\usepackage{multirow}
\usepackage{bm}
\usepackage{xcolor}
\usepackage{url}
\usepackage[hidelinks]{hyperref}
\usepackage{soul}
\usepackage{soulutf8}

\graphicspath{{figures/}}

\newcommand{\ours}{HarvestPoint-ACT}
\newcommand{\hp}{harvest point}
\newcommand{\R}{\mathbb{R}}

\sethlcolor{yellow}

\title{\LARGE \bf
\ours{}: Explicit Target Selection and Harvest-Point Conditioning for Robotic Fruit Harvesting under Occlusion
}

\author{Hanying Hu$^{1,*}$, Weipeng Li$^{1,*}$, Yikun Huang$^{1}$,
Hao Chen$^{1,\dagger}$, Zhengtao Hu$^{2}$, Changcai Yang$^{1}$, and Weiwei Wan$^{3}$%
\thanks{This work was supported by the Fujian Provincial Department of
Education under Grant JAT241022 and by the Fujian Provincial Forestry
Bureau under Grant 2026FKJ3.}%
\thanks{$^{1}$College of Computer and Information Sciences, Fujian Agriculture and
Forestry University, Fuzhou, China.}%
\thanks{$^{2}$School of Mechatronic Engineering and
Automation, Shanghai University, Shanghai, China.}%
\thanks{$^{3}$Graduate School of Engineering Science,
The University of Osaka, Osaka, Japan.}%
\thanks{$^{*}$Hanying Hu and Weipeng Li contributed equally.}%
\thanks{$^{\dagger}$Corresponding author: Hao Chen,
{\tt\small chenhaox@outlook.com}}%
}

\begin{document}

\maketitle
\thispagestyle{empty}
\pagestyle{empty}

% =============================================================================
\begin{abstract}
End-to-end imitation learning avoids hand-made robot motion for approaching and
grasping, but the policy must still decide which fruit to pick and where to
close the gripper. 
Occlusion can make the policy lose the selected fruit during harvesting, and
the correct closing point is difficult to infer from pixels alone. 
This paper presents \ours{}, which makes both decisions explicit in perception
and provides them to the policy. 
An instance segmentation front end with a
keypoint branch predicts a mask and a \hp{} for each visible fruit, where the
\hp{} specifies the location to close the gripper.
A scheduler ranks detected candidates by occlusion and travel distance 
and selects one target. After each attempt, it redetects and reranks the
candidates because the canopy may have changed. The selected fruit
is encoded for an action chunking transformer as an eight-dimensional
state, containing the absolute \hp{}, the vector from the gripper to that point,
a validity flag, and a confidence score. 
When the selected fruit is temporarily
undetected, the system retains the last \hp{} estimate in the robot base frame
and marks it as stale, and aborts the attempt if the loss persists.
On a canopy mock-up, \ours{} achieves a success rate of $88\%$, and of $75\%$
under heavy occlusion.
\end{abstract}

% \begin{keywords}
% Agricultural robotics, imitation learning, action chunking, grasp-point
% detection, instance segmentation, occlusion handling.
% \end{keywords}

% =============================================================================
\section{Introduction}
\label{sec:intro}

Selective fruit harvesting is still one of the least automated operations in
agriculture. The main bottleneck is perception--action coupling:
the system needs to select fruit in cluttered, deformable, and self-occluding
canopies and reach it without damaging the fruit \cite{bac2014harvesting}. 
Despite improvements in cycle time
and success rate, field performance for many crops remains insufficient for
commercial deployment, with occlusion and target ambiguity remaining major
sources of failure \cite{zhou2022intelligent}.

Classical harvesting pipelines decompose the task into detection, 3-D
localization, motion planning, and a servoed final approach
\cite{kang2020visual}. Although this structure is
interpretable, errors propagate across stages:
planning inherits pose-estimation errors, and contact-induced branch deformation
is difficult to encode as a planning constraint. Imitation learning instead
maps observations directly to actions \cite{zhao2023act,chi2023diffusion}, avoiding explicit models of these interactions. 
ACT and related visuomotor policies have
recently been explored for viewpoint planning, clustered strawberry picking,
and open-field pepper harvesting
\cite{li2025viewplanning,fei2025strawberry,kim2025pepper}.

However, applying ACT to harvesting introduces two challenges: which fruit to approach and where to grasp it. 
First, similar-looking fruits make the target identity ambiguous. Occlusion, loss of
the target from the wrist-camera view, or increased salience of a neighboring
fruit may cause the policy to drift or stall \cite{barth2016eyeinhand,li2025viewplanning}.
Second, selecting the correct fruit does not specify a valid grasp point. 
The grasp point is a task-specific affordance constrained by the
peduncle, supporting branch, and gripper geometry
\cite{sa2017peduncle}. It is therefore not equivalent to
a generic image feature such as the mask centroid.
ACT must otherwise learn this metric grasp point implicitly from demonstrations.

\begin{figure}[t]
    \centering
    \includegraphics[width=0.95\columnwidth]{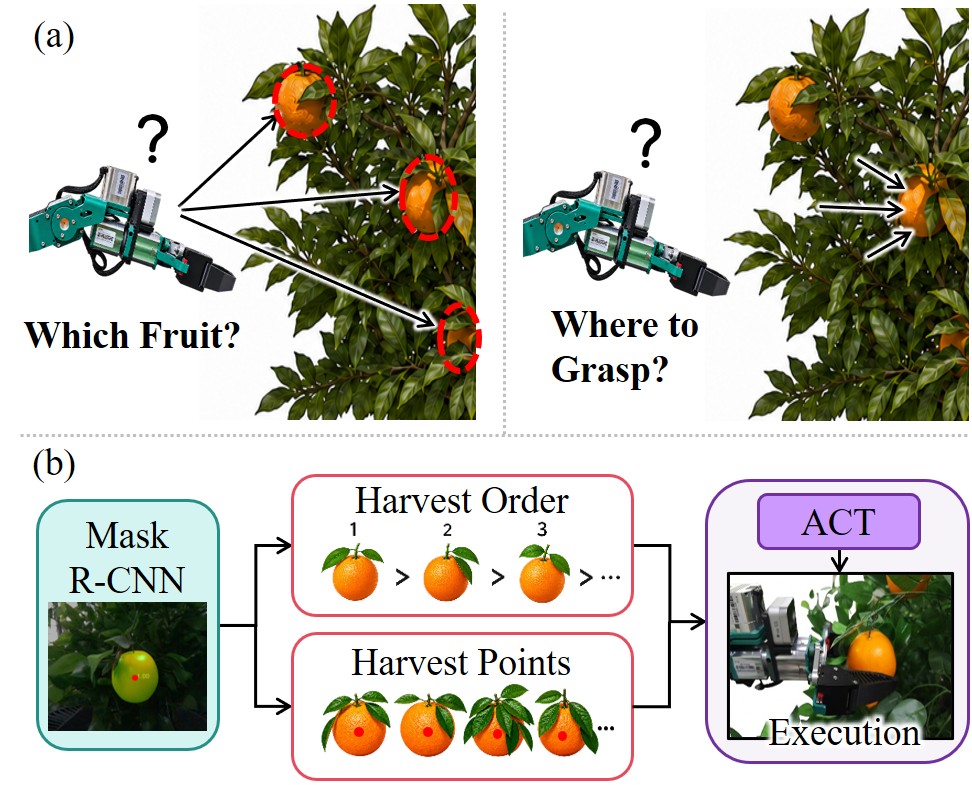}
    \caption{
    (a) The robot must determine which fruit to pick and where to grasp it during harvesting.
    (b) Overview of HarvestPoint-ACT.}
    \label{fig:introduction}
\end{figure}

Both challenges involve information that are already available in the perception output.
An instance segmentation front end supplies a per-fruit mask, a keypoint branch estimates the harvest point
\cite{he2017maskrcnn}.  We therefore develop \ours{}, in which a
scheduler selects the target fruit and an eight-dimensional harvest-point state
conditions ACT on its metric grasp point. A persistence mechanism maintains
this target through temporary detection failures. This design leads to two
contributions:
\begin{itemize}
  \item \textbf{Target-specific \hp{} conditioning for ACT.} We select one fruit
  from the detected instances and condition ACT on its grasp point through an
  eight-dimensional state. The state encodes its grasp point in the robot base frame 
  and relative to the end effector, together with a validity flag and confidence score.
  This state explicitly tells the policy which fruit to grasp and where. The ACT
  backbone is unchanged apart from one input token.
  \item \textbf{An occlusion-aware closed-loop harvesting system.} A scheduler
  ranks fruits by occlusion and travel cost and reranks them after each attempt.
  During temporary detection failures, a persistence rule holds the \hp{} in the
  robot base frame and marks it as stale; persistent failures trigger
  rescheduling.
\end{itemize}

\section{Related Work}
\label{sec:related}

\subsection{Robotic Fruit Harvesting}

Field studies and reviews identify occlusion, clustered fruit, and changes in
canopy geometry as recurring causes of harvesting failure
\cite{arad2020development,zhou2022intelligent}. Most systems use a
modular pipeline: perception estimates a fruit pose or picking point, and a
motion planner and visual servo controller execute the approach
\cite{kang2020visual,barth2016eyeinhand}. We retain the perception front end but
provide its estimated \hp{} to a learned visuomotor policy as an explicit goal.
This keeps the task geometry explicit while allowing the approach behavior to
be learned from demonstrations.

Picking order also affects cycle time and success rate
\cite{kurtser2020planning}. A sequence planned from the initial detections can
become outdated after a pick because nearby fruits may move or become visible.
Our scheduler therefore selects one fruit based on occlusion and travel cost,
then redetects and reranks the candidates after every attempt.

\subsection{Imitation Learning for Harvesting}

Behavior cloning learns visuomotor policies from teleoperated demonstrations
\cite{chi2023diffusion,zhu2022viola}. ACT \cite{zhao2023act}
predicts short sequences of actions
rather than one action at a time, improving temporal consistency during
manipulation. In harvesting,
imitation learning has been used for viewpoint planning, clustered strawberry
picking, and open-field pepper harvesting
\cite{li2025viewplanning,fei2025strawberry,kim2025pepper}. Our focus is instead to
specify which fruit to grasp and where, and to preserve that goal during
temporary detection loss. We retain the ACT backbone and add the
eight-dimensional \hp{} state as one encoder token.

\subsection{Explicit Goal Conditioning for Manipulation}

Conditioning a policy on an explicit goal is common in goal-conditioned
imitation and reinforcement learning \cite{lynch2020play}, where the
goal is usually a target image or a latent plan. Keypoint methods express manipulation goals
more directly as semantically meaningful points. For example, kPAM
\cite{manuelli2019kpam} defines manipulation objectives with keypoint
affordances. Object-centric methods such as VIOLA \cite{zhu2022viola} instead
condition a policy on detected object regions, while visual grasp synthesis
predicts explicit grasp poses \cite{fang2020graspnet}.

\ours{} combines semantic and geometric target information in one \hp{}, which
links ACT to the fruit selected by the scheduler. The state carries both the
manipulation goal and the reliability of its estimate.

\subsection{Target Persistence Under Occlusion}

Video object segmentation uses appearance and temporal memory
to maintain targets through occlusion
\cite{cheng2022xmem,doersch2023tapir}. In dense clusters, however, neighboring
fruits often look alike, so an incorrect visual match changes the policy's
target. We use a conservative matching rule based on mask overlap and 3-D
distance. If no detection matches, the system keeps the last \hp{} in the robot
base frame and marks it as stale. A persistent miss triggers redetection and
rescheduling. This strategy accepts a temporary loss of detection rather than
risking a switch to the wrong fruit.

\section{Method}
\label{sec:method}

\begin{figure*}[t]
  \centering
\includegraphics[width=0.95\textwidth]{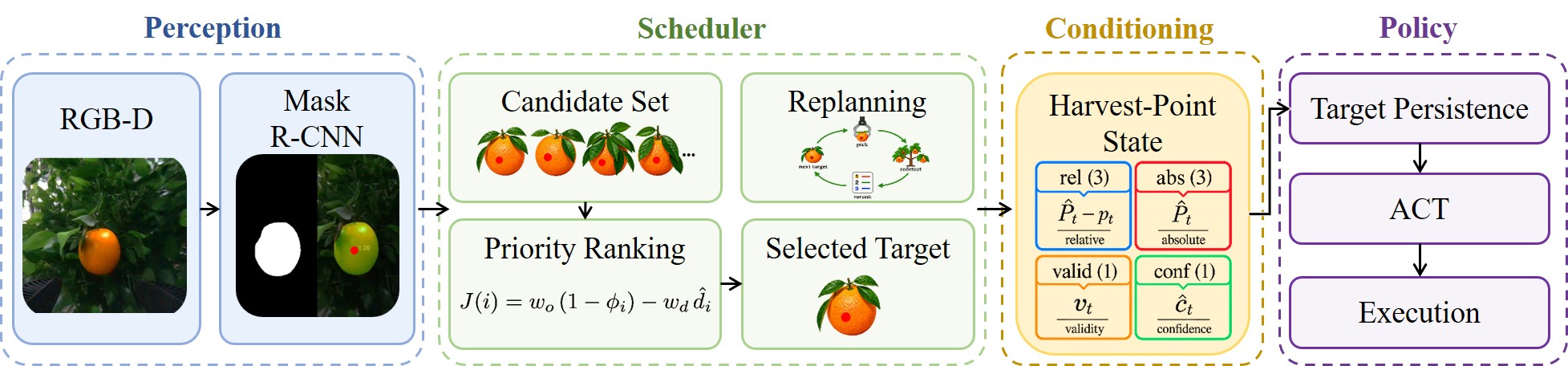}
   \caption{Overview of \ours{}. Instance segmentation yields a mask and a
  \hp{} per fruit. A task-level scheduler selects the target. The target
  enters the policy as an eight-dimensional harvest-point state alongside the RGB
  frame and proprioception, and the transformer regresses a chunk of $H$ future
  end-effector displacements.}
  \label{fig:overview}
\end{figure*}

\subsection{Problem Formulation}
\label{sec:formulation}

We consider a manipulator with an eye-in-hand RGB-D camera in front of a canopy
that contains an unknown number of fruits. A harvesting episode consists of
repeatedly selecting one fruit, approaching it, closing the gripper at a suitable
point on the fruit body, and detaching it.

We separate the problem into a \emph{task level} and a \emph{motion level}. At the
task level, a scheduler observes the current scene, forms the set of harvestable
candidates, and selects one of them. At the motion level, a visuomotor policy
executes the approach and the grasp for the selected target. The policy
receives the identity of the target as an input instead of having to guess it.

At time $t$ the policy receives the observation
$o_t = (I_t^{\text{rgb}}, \bm{x}_t, \bm{e}_t)$. It consists of an RGB image
$I_t^{\text{rgb}} \in \R^{3 \times 480 \times 640}$, an end-effector state
$\bm{x}_t \in \R^{7}$, and a harvest-point state $\bm{e}_t \in \R^{8}$. The
policy outputs a chunk of $H$ future actions
\begin{equation}
  \bm{A}_t = (\bm{a}_t, \bm{a}_{t+1}, \ldots, \bm{a}_{t+H-1})
             \in \R^{H \times 7},
  \label{eq:chunk}
\end{equation}
with $H = 100$. Each action
\begin{equation}
  \bm{a}_\tau = [\Delta x, \Delta y, \Delta z, \Delta r_x, \Delta r_y, \Delta r_z,
g]^\top
\end{equation}
is a relative end-effector displacement expressed in the base frame, with a target
gripper opening $g$. The end-effector state
$\bm{x}_t = [\bm{p}_t^\top, \bm{r}_t^\top, g_t]^\top$ contains the absolute
end-effector position $\bm{p}_t \in \R^3$, its rotation vector
$\bm{r}_t \in \R^3$, and the current gripper opening.

\subsection{Perception and Harvest-Point Extraction}
\label{sec:perception}

Each RGB-D frame $(I_t^{\text{rgb}}, D_t)$ is processed by a Mask R-CNN
\cite{he2017maskrcnn}\footnote{Any instance segmentation model that also predicts
keypoints could take its place; we do not compare detection architectures.} with an additional keypoint branch, fine-tuned on our
own annotated harvesting dataset.\footnote{The dataset is sampled from the RGB
videos recorded during teleoperated demonstration collection. Each annotated
frame carries instance masks and \hp{}s. Dataset size is reported in
Section~\ref{sec:setup}.} It produces a set of detections
\begin{equation}
  \mathcal{D}_t = \{ (M_i, u_i, c_i) \}_{i=1}^{N_t},
  \label{eq:detections}
\end{equation}
where $M_i \in \{0,1\}^{480 \times 640}$ is an instance mask, $u_i$ the predicted
\hp{} in image coordinates, and $c_i \in [0,1]$ a confidence score. Detections
with $c_i < c_{\min}$ are discarded to remove false positives on foliage and background clutter
 from the candidate set.

\begin{figure}[!ht]
  \centering
  \includegraphics[width=0.95\linewidth]{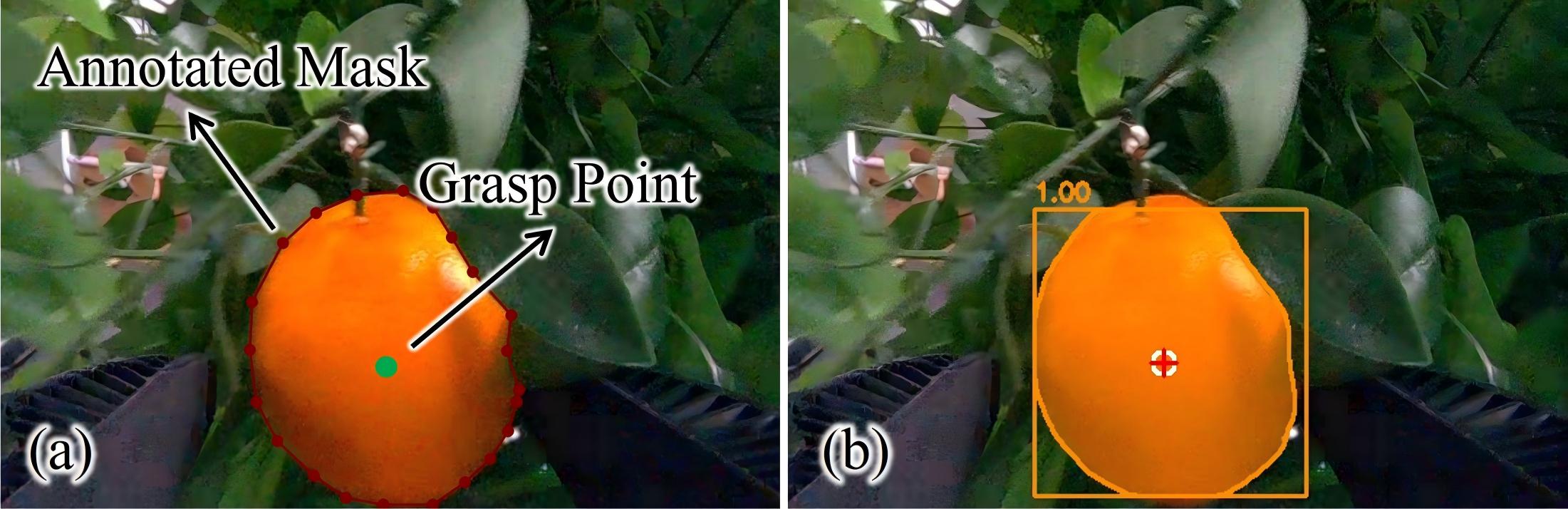}
  \caption{Annotation versus detection on the same wrist-camera views.
  (a) A manually annotated mask and a harvest point.
  (b) Mask R-CNN predictions (masks, keypoints, and confidences).}
  \label{fig:annotation_detection}
\end{figure}

The \hp{} is the location at which the gripper should contact the fruit body. We do
not use the mask centroid: it carries no information about the peduncle, and for a
partially occluded fruit it is biased toward the visible side and can fall outside
the graspable region. We therefore define the \hp{} semantically and label it by
hand. For every fruit instance in the training set, an annotator marks the point
on the fruit body at which the gripper can close without interfering with the
peduncle or with the supporting branch \cite{sa2017peduncle}.
Fig.~\ref{fig:annotation_detection} compares manual annotations with the
detector outputs on the same wrist views.

Depth at a single pixel is not reliable in a canopy. Leaf pixels can enter the
mask, and fruit edges can produce depth artifacts. We therefore check the
predicted point against the median depth $\tilde{z}_i$ inside the mask before
back-projecting it. If the depth at $u_i$ is within a threshold $\tau_z$ of
$\tilde{z}_i$, i.e. $|D_t(u_i) - \tilde{z}_i| \le \tau_z$, we set the pixel $u_i^{\star} = u_i$. 
Otherwise we search a local window $\mathcal{W}(u_i)$ centered at $u_i$,
\begin{equation}
  u_i^{\star} = \arg\max_{u \in \mathcal{W}(u_i) \cap M_i}
        \Big[ \lambda_{d} \, \mathrm{EDT}(u; M_i)
            - \lambda_{z} \, \big| D_t(u) - \tilde{z}_i \big| \Big],
  \label{eq:pickpoint}
\end{equation}
where $\mathrm{EDT}(\cdot; M_i)$ is the Euclidean distance transform of the mask,
and $\lambda_{d},\lambda_{z}>0$ weight the two terms. The first term keeps the
corrected pixel away from the mask boundary, and therefore away from occlusion
edges. The second term rejects pixels whose depth is inconsistent with the fruit
body. If no pixel in the window satisfies the same threshold $\tau_z$, the fruit
is dropped from the candidate set. The retained pixel is
back-projected with the camera intrinsics $K$ and transformed from the camera frame
$\mathcal{C}$ into the robot base frame $\mathcal{B}$,
\begin{equation}
\begin{aligned}
  P_i^{\mathcal{C}}
  &= D_t(u_i^{\star}) \, K^{-1}
     \begin{bmatrix} u_i^{\star} \\ 1 \end{bmatrix}, \\
  \begin{bmatrix} P_i^{\mathcal{B}} \\ 1 \end{bmatrix}
  &= T_{\mathcal{C}}^{\mathcal{B}}(t)
    \begin{bmatrix} P_i^{\mathcal{C}} \\ 1 \end{bmatrix},
\end{aligned}
  \label{eq:backproj}
\end{equation}
where
$T_{\mathcal{C}}^{\mathcal{B}}(t) = T_{\mathcal{E}}^{\mathcal{B}}(t)
T_{\mathcal{C}}^{\mathcal{E}}$ composes the forward kinematics of the arm, which
maps the end-effector frame $\mathcal{E}$ into $\mathcal{B}$, with the hand--eye
calibration.

\subsection{Task-Level Harvest Scheduling}
\label{sec:scheduling}

The candidate set $\mathcal{C}_t$ consists of the detections that survive the
confidence and depth checks of Section~\ref{sec:perception}. The scheduler ranks
them and selects one.

\subsubsection{Occlusion Index}
\begin{figure}[!ht]
  \centering
\includegraphics[width=0.95\linewidth]{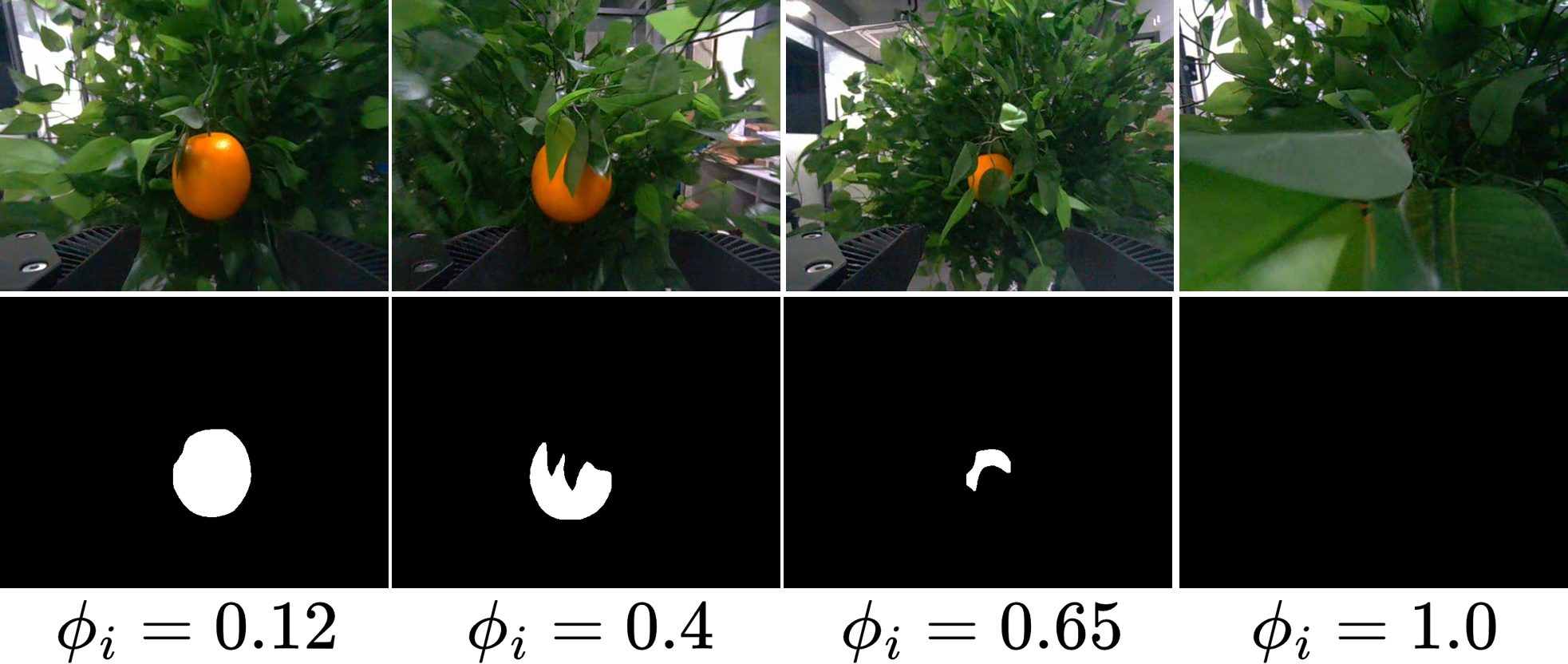}
  \caption{Occlusion index on wrist-camera views.
  For each fruit, the occlusion index $\phi_i$ in
  \eqref{eq:occlusion} measures the fraction of the mask boundary that
  is adjacent to near occluders (neighboring fruits or foliage).
  Low $\phi_i$ indicates a freer fruit that should be picked earlier.
  High $\phi_i$ indicates a deeply occulded fruit.}
  \label{fig:occlusion}
\end{figure}

The occlusion index $\phi_i$ is the fraction of a fruit's mask boundary
$\partial M_i$ that has an occluder beside it. The occluder set $\Omega_i$ contains the pixels of the
other masks $M_j$, $j \neq i$, and the pixels outside $M_i$ with
$D_t(u) < \tilde{z}_i - \delta$, which are the foliage and branches that pass in
front of the fruit. Background pixels farther away than the fruit depth are excluded.
The occlusion index is
\begin{equation}
  \phi_i = \frac{\big| \partial M_i \cap \mathcal{N}_{r}(\Omega_i) \big|}
                {\big| \partial M_i \big|} \in [0,1],
  \label{eq:occlusion}
\end{equation}
where $\mathcal{N}_{r}(\cdot)$ dilates a set by $r$ pixels. The intersection
therefore collects the boundary pixels that lie within $r$ pixels of an
occluder, and $\phi_i$ is the fraction of the boundary occupied by occluders' dilation.
A fruit that hangs free has $\phi_i \approx 0$. A fruit whose outline is mostly bounded by
leaves, or by neighboring fruits in front of it, has $\phi_i$ close to $1$.
Fig.~\ref{fig:occlusion} illustrates the occlusion index on typical
canopy views.
\subsubsection{Priority Score}
Candidates are ranked by
\begin{equation}
  J(i) = w_o \, (1 - \phi_i) - w_d \, \hat{d}_i ,
  \label{eq:score}
\end{equation}
where $\hat{d}_i$ is the travel distance from the current end-effector pose to the
pre-grasp pose, normalized by the workspace diameter. The ordering is
least-occluded-first, so outer fruits are removed before the fruits behind them. This
raises the success probability of the current attempt and reduces the disturbance
to the remaining fruits.

\subsubsection{Replanning After Every Pick}
The scheduler selects $i^\star = \arg\max_{i \in \mathcal{C}_t} J(i)$, executes
one attempt, and then redetects and reranks from scratch. We do not compute a full
harvesting sequence in advance because detaching a
fruit releases the load on a branch and the fruits around it may move. 
Occlusion can also change, and hidden fruits can appear. 

\subsection{Harvest-Point Conditioning}
\label{sec:conditioning}

Once a target $i^\star$ is selected, it is passed to the policy in task
space, as an eight-dimensional harvest-point state
\begin{equation}
  \bm{e}_t = \big[\,
      \underbrace{\hat{P}_t - \bm{p}_t}_{\text{relative } (3)} ;\;
      \underbrace{\hat{P}_t}_{\text{absolute } (3)} ;\;
      \underbrace{v_t}_{\text{validity } (1)} ;\;
      \underbrace{\hat{c}_t}_{\text{confidence } (1)}
    \,\big] \in \R^{8},
  \label{eq:envstate}
\end{equation}
where $\hat{P}_t \in \R^3$ is the current estimate of the \hp{} in the base frame,
$v_t \in \{0,1\}$ indicates whether the estimate was refreshed by a detection at
this time step, and $\hat{c}_t$ is the associated detection confidence. The
relative term is the vector from the gripper to the \hp{}, expressed in the
frame of the actions in \eqref{eq:chunk}. It points the direction the gripper needs
to move. The absolute term describes where that harvesting point is in the workspace, which the
relative vector alone does not say.
The validity and confidence entries tell the policy how much to trust the estimate.
Section~\ref{sec:persistence} describes how the estimate is maintained when detection fails.

The policy receives no mask. Since $\bm{e}_t$ already identifies the selected fruit, adding the
target mask as a fourth image channel gives no further gain
(Section~\ref{sec:experiments}), although masks are still used to compute the
\hp{} and $\phi_i$.

\subsection{Policy, Training, and Inference}
\label{sec:architecture} 

We use ACT \cite{zhao2023act} in the LeRobot implementation
\cite{cadene2026lerobot}. The only change is at the input. The harvest-point
state $\bm{e}_t$ is projected to its own encoder token, next to the token of
the end-effector state $\bm{x}_t$.
The policy is trained by behavior cloning on $100$ teleoperated demonstrations
with the objective of ACT.

At execution, the policy is queried at $12$\,Hz. It predicts a chunk of $H$
actions \eqref{eq:chunk} and executes only the first $k=50$ of them, then
is queried again on a new observation. This receding-horizon execution limits
open-loop drift without querying at every timestep.
The predicted displacements are added to the current pose and the manipulator is commanded to
that pose.
 Fig.~\ref{fig:overview} summarizes the data flow.

\subsection{Target Persistence and Graceful Degradation}
\label{sec:persistence}

During an approach, the detector can lose the selected fruit because a leaf or a
neighboring fruit can occlude it, or the target can leave the wrist camera's
field of view. The harvest-point state \(\bm{e}_t\) must remain defined through
these misses. We maintain it in three steps. 

\subsubsection{Step 1: Match and Update}
At each cycle we match a detection to the selected fruit if its mask overlaps
the previous target mask and its 3-D point is close to $\hat{P}_t$,
\begin{equation}
  \mathrm{IoU}\big(M_j, \hat{M}\big) > \tau_{\text{iou}}
  \quad \text{and} \quad
  \big\| P_j^{\mathcal{B}} - \hat{P}_t \big\|_2 < \tau_{p}.
  \label{eq:assoc}
\end{equation}
where $M_j$ and $P_j^{\mathcal{B}}$ belong to detection $j$ of the current cycle,
$\hat{M}$ is the mask of the target in the previous cycle, $\tau_{\text{iou}}$
is an overlap threshold, and $\tau_{p}$ is a distance threshold. On a match
 we blend the new point into the estimate,
$\hat{P}_t = (1-\alpha)\hat{P}_{t-1} + \alpha P_j^{\mathcal{B}}$ with
$\alpha \in (0,1)$, and set $v_t = 1$.

\subsubsection{Step 2: Hold the Estimate}
When no detection satisfies \eqref{eq:assoc}, we do not search for the fruit in
the image. Instead, we assume the fruit is barely moves during the few seconds of an approach. 
We keep $\hat{P}_t = \hat{P}_{t-1}$, and $\bm{e}_t$ therefore still points at the last known location.
 We set $v_t = 0$ and decay the confidence,
$\hat{c}_t \leftarrow \gamma \hat{c}_{t-1}$ with $\gamma \in (0,1)$, so the
policy knows the estimate is old. 

\subsubsection{Step 3: Abort and Reschedule}
If the miss lasts more than $T_{\max}$ cycles, we stop using the estimate. The arm
retracts and redetects the fruits from scratch. The scheduler of
Section~\ref{sec:scheduling} may select a different fruit.
 The failed attempt is dropped, and the episode continues.

Algorithm~\ref{alg:harvest} shows how these three steps interact with the
scheduler.

\begin{algorithm}[t]
\caption{Harvest loop with scheduling and persistence}
\label{alg:harvest}
\small
\begin{algorithmic}[1]
\While{fruits remain and the time budget is not exceeded}
  \State $\mathcal{C}_t \gets \textsc{Detect}(I_t^{\text{rgb}}, D_t)$
         \Comment{\eqref{eq:detections}--\eqref{eq:backproj}}
  \If{$\mathcal{C}_t = \emptyset$}
    \State \textbf{break}
  \EndIf
  \State $i^\star \gets \arg\max_{i \in \mathcal{C}_t} J(i)$
         \Comment{\eqref{eq:score}}
  \State $\hat{P}_t \gets P_{i^\star}^{\mathcal{B}}$;\quad
         $\hat{M} \gets M_{i^\star}$;\quad
         $n_{\text{lost}} \gets 0$
  \While{the grasp is not completed}
    \State $\mathcal{C}_t \gets \textsc{Detect}(I_t^{\text{rgb}}, D_t)$
    \If{a detection $j$ satisfies \eqref{eq:assoc}}
      \State $\hat{P}_t \gets (1-\alpha)\hat{P}_t + \alpha P_j^{\mathcal{B}}$
      \State $v_t \gets 1$;\; $\hat{c}_t \gets c_j$;\;
             $\hat{M} \gets M_j$;\; $n_{\text{lost}} \gets 0$
             \Comment{Step 1}
    \Else
      \State keep $\hat{P}_t$;\; $v_t \gets 0$;\;
             $\hat{c}_t \gets \gamma \hat{c}_t$
      \State $n_{\text{lost}} \gets n_{\text{lost}} + 1$
             \Comment{Step 2}
      \If{$n_{\text{lost}} > T_{\max}$}
        \State retract; \textbf{break}
        \Comment{Step 3}
      \EndIf
    \EndIf
    \State form $o_t$ by \eqref{eq:envstate}
    \State $\hat{\bm{A}}_t \gets \pi_\theta(o_t)$;\;
           execute the first $50$ actions
  \EndWhile
  \If{the grasp was completed}
    \State retract and release the fruit
  \EndIf
\EndWhile
\end{algorithmic}
\end{algorithm}

\section{Experiments}
\label{sec:experiments}

\subsection{Experimental Setup}
\label{sec:setup}
\begin{figure}[ht]
\centering
    \begin{minipage}[c]{0.48\linewidth}
        \includegraphics[width=\linewidth]{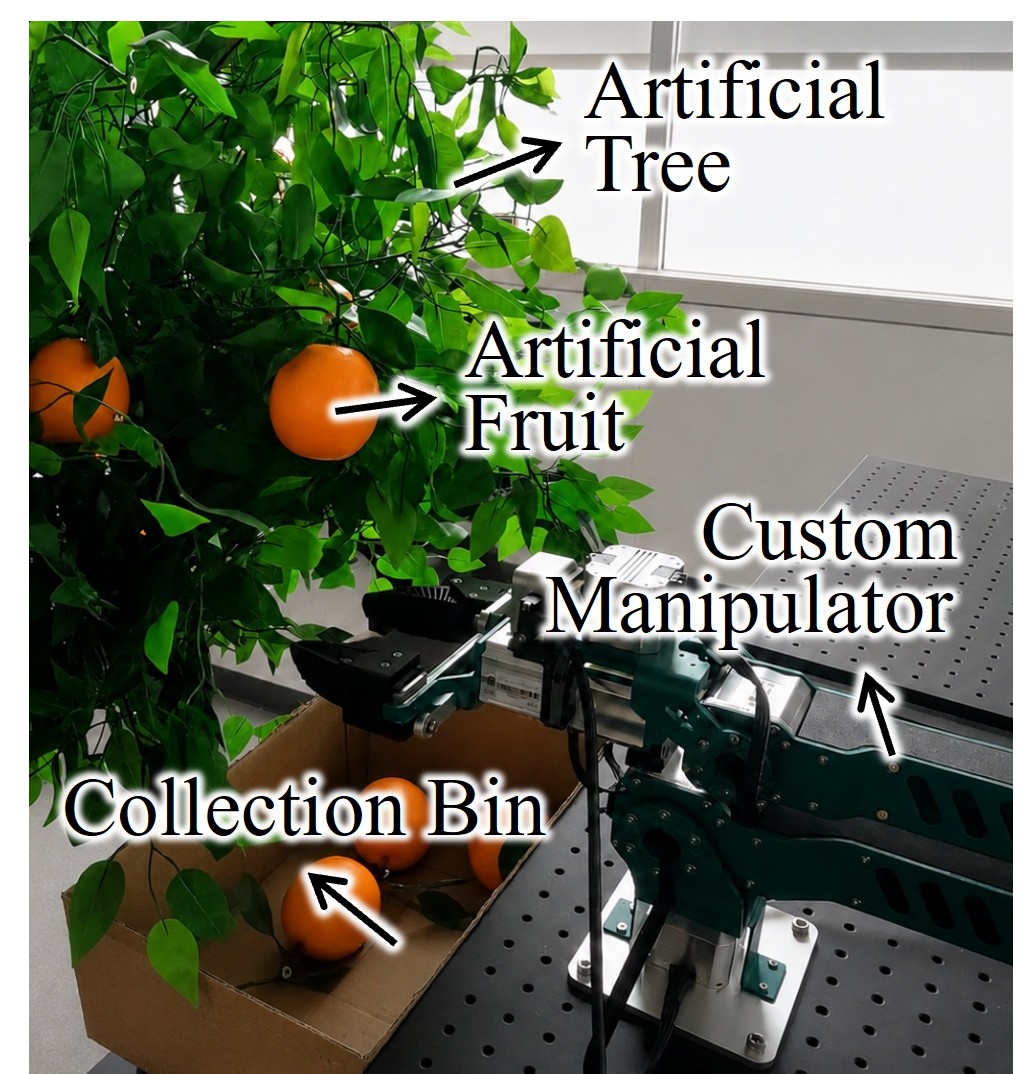}
    \end{minipage}% 
    \hspace{0.01\linewidth}
    \begin{minipage}[c]{0.47\linewidth}
        \caption{Experimental setup for robotic fruit harvesting. The setup consists of the custom-design manipulator with an Intel RealSense D405
mounted on the wrist, an artificial tree with artificial oranges, and a collection bin.}
        \label{fig:experimental_setup}
    \end{minipage}
\end{figure}

Fig.~\ref{fig:experimental_setup} shows the setup. All experiments use a
custom-designed six-degree-of-freedom manipulator with an Intel RealSense D405
mounted on the wrist. Perception
and the policy run on a workstation with an AMD Ryzen~7 9800X3D CPU, $48$\,GB of
RAM, and an NVIDIA RTX~3090 GPU. The code is written in Python. Mask R-CNN uses
PyTorch, and the policy is trained with LeRobot. \textbf{The attached video shows \ours{} harvesting on the real robot}.

The scene is an artificial citrus tree in a laboratory. Artificial oranges hang
from its branches, and a collection bin stands beside the robot base. Each
orange is held by a magnet, so the force needed to pull a fruit off is the same
for every fruit and every trial.

We use oranges because their color contrasts strongly
with the leaves, so segmentation is not the limiting factor and the comparison in
Section~\ref{sec:baselines} reflects the policies rather than the detector.
A crop whose peduncle must be cut would need a different end effector and a different harvest point.
 
Demonstrations are collected by teleoperation, using a leader arm of the same
design as the manipulator. A \emph{layout} is one arrangement of the oranges on
the tree, and we rehang the fruits between layouts to vary occlusion and reach.
We record $100$ demonstrations, each covering one attempt on a
single fruit. From
the recorded RGB streams we annotate $693$ frames, and use them to fine-tune the Mask R-CNN front end.

% \begin{figure}[t]
%     \centering
%     \includegraphics[width=\columnwidth]{figs/exp_env.jpg}
%     \caption{Experimental setup for robotic fruit harvesting. The setup consists of the Fafu\_arm robotic platform, an artificial citrus tree with artificial oranges, and a collection basket.}
%     \label{fig:experimental_setup}
% \end{figure}

\subsection{Evaluation Metric}
Every method is evaluated on the same $4$ fruit layouts. On each layout
the robot runs the loop of Algorithm~\ref{alg:harvest} until the candidate set
is empty.

We report three metrics.
1) \emph{Success} is the fraction of attempts in which the gripper closes on the
body of the scheduled fruit, separates it from the branch, and releases it into
the collection bin. A fruit that is knocked off the tree by contact instead of being
grasped counts as a failure. Artificial fruit cannot be scored for bruising, so
we do not report fruit damage.
2) \emph{Approach success} is the fraction of attempts in which the gripper reaches
the selected fruit. It separates a failure of the approach from a failure of the
grasp close.
3) \emph{Cycle time} is how long one attempt takes, from the start of the detection to the release of the fruit. We average it over successful attempts only, so
it is undefined for a method that never succeeds.
Closing on a fruit other than the selected one counts as a failure for both
success and approach success.

\subsection{Comparison With Baselines}
\label{sec:baselines}
We compare \ours{} with four baselines:
\begin{enumerate}
    \item \emph{ACT (RGB only)} is the original formulation. It sees the
RGB frame and $\bm{x}_t$ and nothing else, so it has to infer from the image
both which fruit to pick and where to close the gripper.
    \item \emph{ACT + masks} adds fruit masks as a fourth image channel.
The policy then sees where the fruits are, but not which one to pick. It therefore tests
whether marking every fruit is enough.
    \item \emph{Diffusion Policy} \cite{chi2023diffusion} keeps the same RGB observation and
replaces the action decoder with a diffusion model.
    \item \emph{SmolVLA} \cite{shukor2025smolvla} is a pretrained vision--language--action
model, fine-tuned on the same demonstrations. Its language channel specifies the
task semantically, but says nothing about where to close the gripper.
\end{enumerate}

All learned methods are trained on the same $100$ demonstrations for the same
number of gradient steps, and are queried at the same rate at execution. 
Every method is evaluated on the same $25$ harvesting cases, which differ in the
position of the fruit and in the surrounding clutter. One case is one
\emph{attempt}. The attempt ends when the fruit is released into the
collection bin, or when the time budget of 10\,s runs out. An attempt that reaches
the fruit but does not detach it before that limit counts as a failure.

\begin{table}[!t]
  \centering
  \caption{Harvesting performance on the canopy mock-up}
  \label{tab:main}
  \begin{tabular}{lccc}
    \toprule
      & Success & Approach & Cycle time \\
       Method    & (\%)    & Success (\%) & (s) \\
    \midrule
    ACT (RGB only) \cite{zhao2023act}        & 0  & 76 & --- \\
    ACT + masks                              & 0  & 80 & --- \\
    Diffusion Policy \cite{chi2023diffusion} & 0  & 80 & --- \\
    SmolVLA                                  & 0  & 84 & --- \\
    \textbf{\ours{}}                         & \textbf{88} & \textbf{92} & 21.5 \\
    \bottomrule
  \end{tabular}
\end{table}

In Table~\ref{tab:main}, all four baselines reach vicinity of the fruit in most
attempts, with approach success between $76\%$ and $84\%$. The approach motion is
therefore learned from the demonstrations even without explicit conditioning.
None of the four grasps a single fruit, and the failure is always the same: the gripper
closes before the fingers touch the fruit. We attribute this to the absence of a distance cue.
The baselines receive the wrist camera image and their own end-effector state, which says
where the gripper is but not where the fruit is. Depth never reaches the policy,
so no input gives the distance to the fruit. Fruit can also be
partly occluded by a leaf or leave the camera view, so the image becomes
ambiguous and small errors take the arm into views that the demonstrations never
cover. 
The harvest-point state supplies that cue, and \ours{} reaches $92\%$ approach
success and $88\%$ grasp success on the same layouts. 

\subsection{Robustness to Occlusion}
Table~\ref{tab:occlusion} reports HarvestPoint-ACT under four occlusion
levels defined by $\phi_i$.
Success rate decreases monotonically as occlusion increases, from
$88.5\%$ for $\phi_i \in [0,0.25)$ to $76\%$ for
$\phi_i \in [0.75,1]$.
The average pick time roughly doubles. The lightly occluded
fruits are completed in about $12$--$13\,\mathrm{s}$, 
whereas heavily occluded fruits take about
$23$--$29\,\mathrm{s}$.

These two trends indicates that dense occluders slow the approach rather
than break it. The arm has to work around leaves, but the \hp{} stays usable.
This is consistent with the persistence rule of Section~\ref{sec:persistence}: when an
occluder removes the detection, $\hat{P}_t$ is still held and
keeps pointing at the fruit.
Overall, explicit harvest-point conditioning remains effective across
occlusion levels. Even in the heaviest level, $76\%$ of fruits are still detached.

\begin{table}[!t]
  \centering
  \caption{Occlusion experiment by occlusion index $\phi_i$
  }
  \label{tab:occlusion}
  \begin{tabular}{lcc}
    \toprule
    Occlusion level & Success (\%) & Avg. (s) \\
    \midrule
    $\phi_i \in [0,\,0.25)$     & 90 & 11.6 \\
    $\phi_i \in [0.25,\,0.50)$  & 80 & 12.3 \\
    $\phi_i \in [0.50,\,0.75)$  & 80 & 23.3 \\
    $\phi_i \in [0.75,\,1.00]$  & 75 & 28.9 \\
    \bottomrule
  \end{tabular}
\end{table} 

\subsection{Failure Modes}
\begin{figure}[t]
    \centering
    \includegraphics[width=0.9\columnwidth]{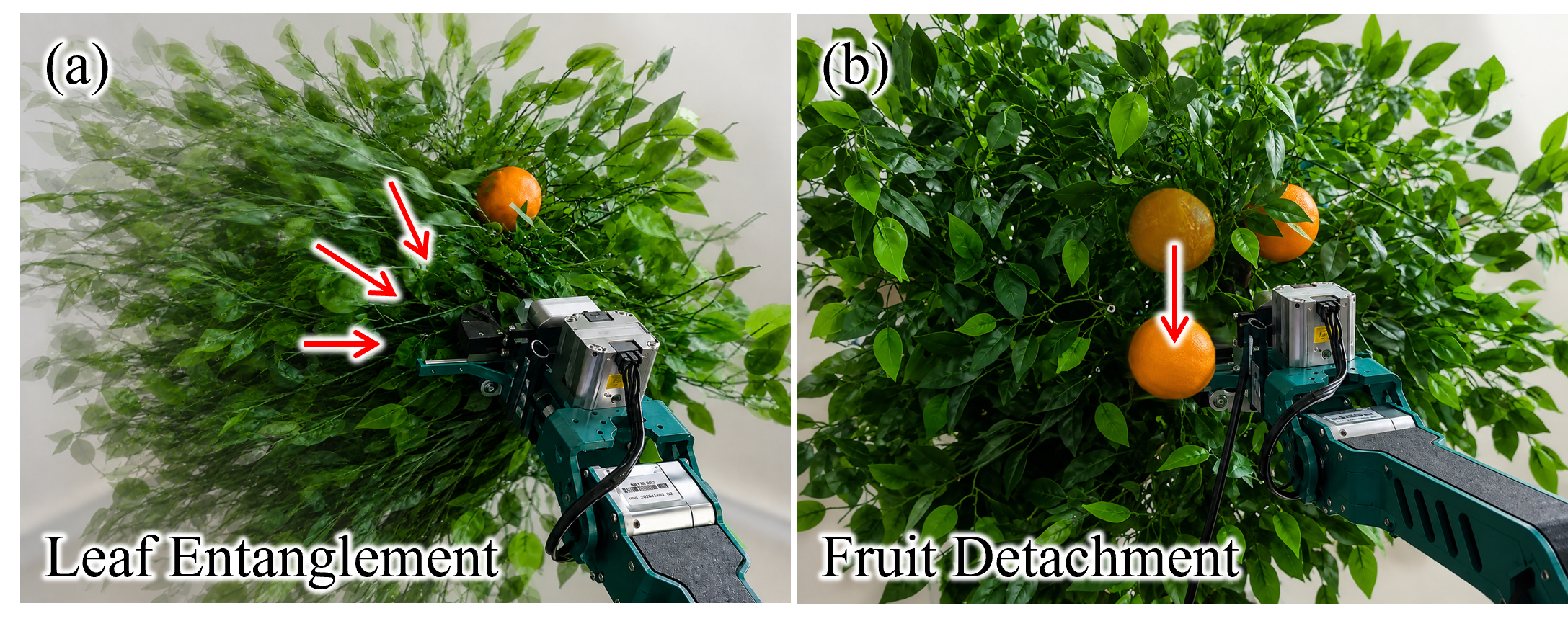}
    \caption{Representative failure cases during robotic fruit harvesting.
    (a) Leaf dragging caused by unintended interaction with surrounding
    foliage. (b) The fruit falls off due to contact during harvesting.}
    \label{fig:failure_cases}
\end{figure}

Two representative failure cases observed during harvesting are
shown in Fig.~\ref{fig:failure_cases}. First, the gripper catches a leaf on the way in and drags it. This
damages the foliage and can bring the arm to a stop before it reaches the fruit. Second, contact knocks the fruit off its magnet before the fingers close, so the attempt ends without a grasp.

Both cases come from contact with the canopy, which the policy does not model. We
assume that the fruit barely moves while the arm approaches, and contact can
violate this. We also do not model entanglement, where the gripper or the fruit
catches on leaves and branches. Sensing such contact and recovering from it is
left to future work.

\section{Conclusion}
\label{sec:conclusion}

We presented \ours{}, a fruit harvesting policy that is told which fruit to pick
and where to close the gripper, instead of having to infer both from pixels.
Instance segmentation predicts a mask and a metric \hp{} per fruit, a scheduler
selects one target and replans after each attempt, and the target enters an action
chunking transformer as an eight-dimensional state. Because the \hp{} is held in
the robot base frame, losing the detection during an approach does not end the
attempt. On a canopy mock-up, \ours{} achieves a success rate of $88\%$, and of $75\%$
under the heaviest occlusion level.

In the future, we will evaluate the system in an orchard, model contact during
harvesting, and integrate a mobile base to reach fruits that a single viewpoint
can neither see nor reach.

% =============================================================================
\bibliographystyle{IEEEtran}
{\let\footnotesize\scriptsize
\bibliography{refs}}

\end{document}